\documentclass{article}
\usepackage{microtype}
\usepackage{graphicx}
\usepackage{booktabs} 
\usepackage{hyperref}

\usepackage[accepted]{icml2024}

\usepackage{amsmath}
\usepackage{amssymb}
\usepackage{mathtools}
\usepackage{amsthm}
\usepackage{xcolor}
\usepackage{authblk}
\usepackage[most]{tcolorbox}
\tcbset{on line, 
        boxsep=2pt, left=0pt,right=0pt,top=2pt,bottom=2pt,
        colframe=white, 
        highlight math style={enhanced}
        }

\usepackage[font=small]{caption}
\usepackage{subcaption}

\usepackage[capitalize,noabbrev]{cleveref}

\theoremstyle{plain}

\theoremstyle{definition}

\theoremstyle{remark}

\usepackage{bookmark}
\usepackage{lipsum}

\usepackage[normalem]{ulem} 
\usepackage{xcolor}

\def\footurl#1{\footnote{\href{#1}{#1}}}

\newcommand\csquare[1][black]{\textcolor{#1}{\,\rule[-0.1em]{0.8em}{0.8em}\,}}

\usepackage[textsize=tiny]{todonotes}

\usepackage{listings}
\definecolor{codegreen}{rgb}{0,0.6,0}
\definecolor{codegray}{rgb}{0.5,0.5,0.5}
\definecolor{codepurple}{rgb}{0.58,0,0.82}
\definecolor{backcolour}{rgb}{0.95,0.95,0.92}

\lstdefinestyle{mystyle}{
    backgroundcolor=\color{white},   
    commentstyle=\color{codegreen},
    keywordstyle=\color{magenta},
    numberstyle=\tiny\color{codegray},
    stringstyle=\color{codepurple},
    basicstyle=\ttfamily\footnotesize,
    breakatwhitespace=false,         
    breaklines=true,                 
    captionpos=b,                    
    keepspaces=true,                 
    numbers=left,                    
    numbersep=5pt,                  
    showspaces=false,                
    showstringspaces=false,
    showtabs=false,                  
    tabsize=2,
    xleftmargin=1.5em,
    framexleftmargin=1.5em
}

\icmltitlerunning{Length-based Overfitting of Elementary Functions in Transfomers}

\begin{document}

\twocolumn[
\icmltitle{Adversarial Testing as a Tool for Interpretability: \\ Length-based Overfitting of Elementary Functions in Transformers
}


\begin{icmlauthorlist}
\icmlauthor{Patrik Zavoral}{mff}
\icmlauthor{Du\v{s}an Vari\v{s}}{mff}
\icmlauthor{Ond\v{r}ej Bojar}{mff}
\end{icmlauthorlist}

\icmlaffiliation{mff}{Department of Mathematics and Physics, Charles University, Prague, Czech Republic}

\icmlcorrespondingauthor{Patrik Zavoral}{patrik.zavoral@gmail.com}

\icmlkeywords{Transformer,Length,OOD,Adversary,Tracr,RASP,Interpretation,Overfitting,Generalization,ICML}

\vskip 0.3in
]



\printAffiliationsAndNotice{}  

\begin{abstract}

The Transformer model has a tendency to overfit various aspects of the training data,
such as the overall sequence length.
We study elementary string edit functions 
using a defined set of error indicators to interpret the behaviour of the 
sequence-to-sequence Transformer.
We show that generalization to shorter sequences is often possible, but confirm that
longer sequences are highly problematic, although partially correct
answers are often obtained. Additionally, we find that other structural characteristics
of the sequences,
such as subsegment length, may be equally important. We hypothesize that the
models learn algorithmic aspects of the tasks simultaneously with structural
aspects but adhering to the structural aspects is unfortunately often preferred
by Transformer when they come into conflict.

\end{abstract}

\section{Introduction}

Transformer \cite{vaswani} has become a widely used architecture in 
Natural Language Processing (NLP) and beyond. It reached state-of-the-art performance in
sequence-to-sequence tasks such as natural language translation,
surpassing recurrent architectures based e.g. on LSTM \cite{lstm}.
Using the Generative Pretrained Transformer (GPT) 
for language modelling
\cite{gpt}
has also become widespread, with continuing efforts for further improvement of
the architecture.
Of particular interest is the extent to which Transformer models
``understand'' and are able to generalize to new, out-of-distribution
scenarios \cite{parrots},
and the extent to which this ability comes from the Transformer architecture itself 
or the sheer volume
of training data combined with exploitation of the similarities between
the training and validation data \cite{overfitting}. 

We study length-based generalization, whereby the novel out-of-distribution
condition is induced solely
by controlling the range of the sequences in the training and validation sets.
This type of generalization is especially apparent in tasks where the pattern is 
elementary, and therefore easily identifiable by humans.
For example, when we illustrate the operation of string reversal on short strings,
humans will correctly reverse also a long string.
Such elementary string edit functions
thus highlight the extent to which universal approximators may be limited
by data.
The elementary functions we experiment with are solvable using very small Transformers (1-2 layers, 1 attention head; 
\citealp{thinking}) and
it is possible to construct
such Transformers implementing these algorithms
written in the Restricted Access Sequence Processing Language (RASP).
We are going in the opposite direction,
with independently trained minimal Transformers, and observe Transformers
fail in learning the generalization.
Interpreting the behaviour of these trained Transformers
then truly forces us to \textit{think like Transformers}, 
instead of forcing Transformers to \textit{think like us} \cite{thinking}.

\subsection{Length-based overfitting}

Humans intuitively generalize over sequence length.
For example, given only a few examples of string reversal on short strings,
humans will correctly reverse long strings too.
This type of generalization however does not quite happen in Transformers.
In both 
machine translation (MT) and 
elementary string editing tasks such as string copying or reversing, the Transformer model
was found to overfit with regard to the training target-side length distribution
\cite{overfitting}.
A simple copy-like subtask (named entity transcription) within the broader task of general machine translation
\label{sec:named-entity-transcription}
fails for large named entity lengths,
despite the overall sentence length fits the training data limit. Moreover,
such performance drop may occur whenever
there is a lack of training data of a specific target-side length range,
shorter or longer \cite{varis}.

\subsection{Elementary functions}

We choose to study \textit{elementary} string editing functions on binary sequences.
This offers a complete control over the training and validation
data, as they can be algorithmically generated.
As previously argued, the elementary functions highlight the extent to which universal approximators are limited
by data, since the pattern easily identified and extrapolated by human agents.
In particular, we will focus on the functions
\textbf{copy}, \textbf{flip}, and \textbf{reverse} 
(identical copy, swap $a$ for $b$ and vice versa, and copy backwards, respectively)
on
$(a|b)^*$ sequences.
These functions are solvable using very small Transformers 
(1-2 layers, 1 attention head), obtained e.g. by
the Tracr compiler from RASP, see \cref{fig:rasp}.

\subsection{RASP}

The Restricted Access Sequence Processing Language (RASP) is a computational
model for Transformer, which captures the unique information flow constraints
of the architecture \cite{thinking}, similarly to previous conceptualizations of RNNs as 
finite state automata \cite{rnn_dfa}. It allows to program a desired logic in a way
that corresponds to an abstraction of the Transformer model, therefore putting
programmers in the Transformer's position and asking them
in some way to
\textit{think like a Transformer} \cite{thinking}.

The input of a RASP program is in two sequences: \texttt{tokens}, encoding 
the user-provided input, and \texttt{indices}, encoding the respective index
range $0, 1, \dots, n-1$, with additional total \texttt{length} \cite{thinking}.
The program consists of a number of pairing and mapping operations on
these sequences.
These operations then roughly
correspond to the layers in the final compiled Transformer \cite{tracr},
resulting in very shallow architectures for the elementary functions 
(\cref{fig:rasp}).

\citet{tracr} introduced Tracr\footurl{https://github.com/google-deepmind/tracr},
a compiler for converting RASP programs into Transformer weights in Haiku.\footurl{https://github.com/google-deepmind/dm-haiku}
The compilation is achieved by encoding the predicate-based pairings
into attention matrices, 
generating matrix lookup tables
for arbitrary functions with a finite domain, 
approximating arbitrary continuous functions to a required accuracy \cite{uni_apx_thm},
and various other mechanisms \cite{tracr}.

\citet{thinking} also compare the attention defined by RASP programs and 
Transformer models trained on the respective tasks, finding
a varying degree of similarity,
with \textbf{reverse} attaining only a partially matching pattern.
Further, \citet{thinking} claim their RASP programs provide a theoretical
upper bounds for the number of heads
and layers needed for the considered tasks.


\begin{figure}

\colorlet{rasp}{teal!100}

\lstset{language=Python,
        escapechar=$,
        emph={length,tokens,indices,EQ},
        emphstyle=\color{rasp},
    }

\begin{lstlisting}
# generates an off-diagonal attention
opp_index = length - indices - 1
flip = $\color{rasp}select$(indices, opp_index, EQ)
reverse = $\color{rasp}aggregate$(flip, tokens)
\end{lstlisting}

\begin{lstlisting}
# encodes the lookup table
f = lambda x: x  
copy = $\color{rasp}map$(f, tokens)
\end{lstlisting}

\begin{lstlisting}[language=Python]
# encodes the lookup table
g = lambda x: {'a': 'b', 'b': 'a'}[x]  
flip = $\color{rasp}map$(g, tokens)
\end{lstlisting}

\caption{Python-style RASP programs for string \textbf{reverse}, \textbf{copy},
and \textbf{flip} 
(reverse a string, identical copy, and swap $a$ for $b$ and vice versa, respectively)
}
\label{fig:rasp}

\end{figure}




\subsection{Adversarial testing and interpretability}

The out-of-distribution performance of neural models is problematic and exploitable
by adversarial attacks. 
\citet{splits} argue for using adversarial train-test splits to counter the overly
optimistic performance estimates of both standard and random splits;
either through maximizing their divergence or by heuristics
such as sequence length. 
An adversary with malicious intent may even present input data that e.g. mobilises certain
spurious training set characteristics. The result looks trivial to humans but confuses
the model.
Different from \citet{splits}, we induce an adversarial domain shift to analyze and 
interpret the architectural biases leading to the performance drop. 
This can lead to a better understanding and
help design improved architectures or training schemes \cite{overfitting}.

The main source of error identified in MT tasks
with a restricted training length range is the tendency of the Transformer models
to condense or stretch their translation hypotheses to match the training
data length range. 
The models
are unlikely to emit the \texttt{EOS} token at positions where \texttt{EOS} was not seen during
training \cite{varis}.
The metrics used in the adversarial length experiments were either
BLEU \cite{bleu} or verbatim accuracy \cite{varis,overfitting}. 
Both can only detect the presence of a
generalization error but don't allow its further analysis. Moreover, BLEU
relaxes the string-matching test in a non-useful way,
as it is based on matching n-grams.
We improve the analysis by defining various special error indicators on the
validation data and tracking them
throughout the training.
This allows us to study the behaviour of the Transformer models in more detail,
and compared to qualitative evaluation of individual hypotheses or 
attention matrices, the indicators
can be robustly aggregated over the entire validation set in a straightforward
way in order to yield quantitative results.

\begin{figure}[t]
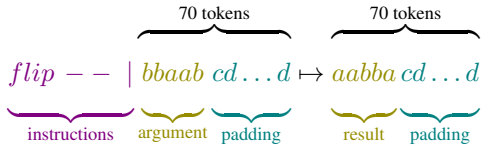

\begin{align*}
    \color{violet}
    \underbrace{\vphantom{\left(\frac{a^{0.3}}{b}\right)} flip\, - - \;\;|}_\text{instructions}
    \color{black}
    \overbrace{
    \color{olive}
    \underbrace{\vphantom{\left(\frac{a^{0.3}}{b}\right)} b b a a b}_\text{argument} 
    \color{teal}
    \underbrace{\vphantom{\left(\frac{a^{0.3}}{b}\right)} c d \dots d}_\text{padding}
    }^\text{70 tokens}
    \color{black}
    \mapsto
    \overbrace{
    \color{olive}
    \underbrace{\vphantom{\left(\frac{a^{0.3}}{b}\right)} a a b b a}_\text{result} 
    \color{teal}
    \underbrace{\vphantom{\left(\frac{a^{0.3}}{b}\right)} c d \dots d}_\text{padding}
    }^\text{70 tokens}
\end{align*}

\caption{Source - target example for \textbf{padded flip}.}
\label{fig:example}
\end{figure}

\section{Experiments}

\subsection{Training}

We use Fairseq\footurl{https://github.com/facebookresearch/fairseq}
for sequence-to-sequence learning \cite{fairseq}.
The $(a|b)^*$ input arguments were generated randomly with their length distributed
evenly within the desired range. 
The target sequences were generated to illustrate three simple functions:
\textbf{copy}, \textbf{flip}, and \textbf{reverse} 
(identical copy, swap $a$ for $b$ and vice versa, and copy backwards, respectively) 
There are possible variations on the elementary tasks. For example, it may be happening
within a more general linguistic task, such as 
named entity transcription in \citet{varis}. 
We 
limit ourselves only to the bare \textbf{simple} variant of each task and to
a \textbf{padded} variant, where the \textbf{simple} source-target pairs were additionally 
padded by
$cd^*$ to the length of 70. \
We introduced this custom and ``visible'' padding to see if the 
premature stopping is due to the actual end of the many training 
sequences or due to the bad estimates of the probability of some 
end-of-sequence token (the standard \texttt{EOS} or the custom $c$).
See \cref{fig:example} for a source-target example
for \textbf{padded flip}. Other tasks follow the same format.
The models were trained only on argument lengths restricted to 
$(20, 30]$ or $(30, 40]$ for up to 400 epochs.
The data in the training and testing sets are independent.
To replicate \citet{overfitting}, the training set size is
28k pairs. The test set size is 2k pairs per every $(k, k+10]$ interval.
Each model was trained either to be fully specialized on one chosen task and variant
(e.g. \textbf{padded flip}), or trained on the concatenated
training data of all tasks of the same \textbf{simple}/\textbf{padded} variant and then evaluated separately on the 
individual tasks. For brevity, e.g. \textbf{flip/all - padded} stands for 
an experiment where we are testing the (\textbf{padded}) \textbf{flip}
task performance of a model trained on all tasks in the \textbf{padded} variant.
For model architecture and training details, see \cref{app:models}.

\begin{figure}[t] 
\begin{align*}
    \overbrace{
    \color{olive}
    \underbrace{\vphantom{\left(\frac{a^{0.3}}{b}\right)} a a b b a}_R 
    \color{teal}
    \underbrace{\vphantom{\left(\frac{a^{0.3}}{b}\right)} c d \dots d}_P
    }^\text{Reference}
    \hspace{1cm}
    \overbrace{
    \color{olive}
    \underbrace{\vphantom{\left(\frac{a^{0.3}}{b}\right)} a a \color{purple}\sout{a}\color{olive} b a}_{\tilde R}
    \color{teal}
    \underbrace{\vphantom{\left(\frac{a^{0.3}}{b}\right)} c \color{purple}\sout{c}\color{teal} d \dots d}_{\tilde P}
    }^\text{Hypothesis}
\end{align*}
    \caption{Reference - hypothesis target example for \textbf{padded flip}
    with two errors marked in purple and underlined.}
    \label{fig:refhyp}
\end{figure}

\begin{figure*}[h]
    \begin{subfigure}[b]{0.5\textwidth}
        \includegraphics[width=\textwidth]{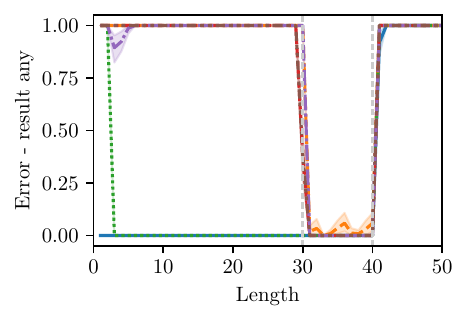}
    \end{subfigure}
    \begin{subfigure}[b]{0.5\textwidth}
        \includegraphics[width=\textwidth]{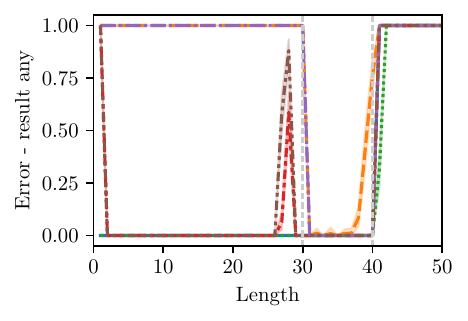}
    \end{subfigure}

    \begin{subfigure}[b]{0.33\textwidth}
        \includegraphics[width=\textwidth]{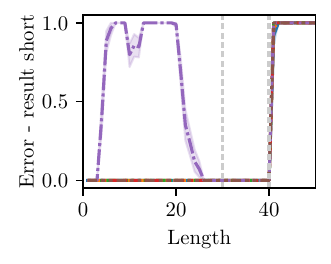}
    \end{subfigure}
    \begin{subfigure}[b]{0.33\textwidth}
        \includegraphics[width=\textwidth]{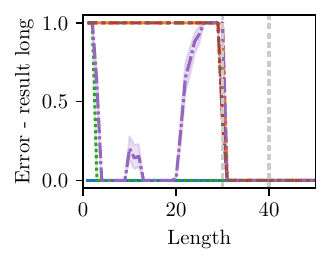}
    \end{subfigure}
    \begin{subfigure}[b]{0.33\textwidth}
        \includegraphics[width=\textwidth]{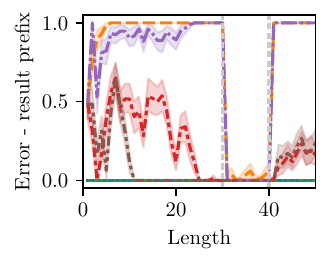}
    \end{subfigure}

    \caption{Selected indicators after 400 epochs. 
        Top row: left - \textbf{simple} tasks, right - \textbf{padded} tasks.
        Bottom row - \textbf{simple} tasks only. \\
        \textbf{Copy} (\csquare[blue!60] blue), 
        \textbf{flip} (\csquare[green!60] green),
        \textbf{reverse} (\csquare[orange!60] orange),
        \textbf{copy/all} (\csquare[red!60] red), 
        \textbf{flip/all} (\csquare[brown!60] brown),
        \textbf{reverse/all} (\csquare[violet!60] violet). \\
        Dashed vertical lines mark the training length range $(30, 40]$.
        Best viewed in color.
    }
    \label{fig:error-any}
\end{figure*}

\begin{table}[t]
    \centering
    \caption{Our error predicates checking the match of reference and hypothesis (marked with tilda) in the (R)esult or (P)adding part.
    Equality with regular expressions on one or both sides is understood as 
    satisfiability, i.e. there exists a satisfying substitution of strings
for the regular expressions such that the equality holds.}
    \label{tab:predicates}

    \begin{tabular}{l l}
        \toprule
        Name & Definition \\
        \midrule
        \textbf{Result} & \\
        \hspace{2mm} any & $\tilde R \ne R$ \\
        \hspace{2mm} short & $|\tilde R| < |R|$ \\
        \hspace{2mm} long & $|\tilde R| > |R|$ \\
        \hspace{2mm} prefix & $\tilde R .^* \ne R .^*$ \\
        \bottomrule
    \end{tabular}
    \begin{tabular}{l l}
        \toprule
        Name & Definition \\
        \midrule
        \textbf{Padding} & \\
        \hspace{2mm} pattern & $\tilde P \ne cd^*$ \\
        \hspace{2mm} short & $|\tilde R \tilde P| < |RP|$ \\
        \hspace{2mm} long & $|\tilde R \tilde P| > |RP|$ \\
        \vphantom{a} & \vphantom{$|\tilde A^*|$} \\
        \bottomrule
    \end{tabular}
\end{table}

\subsection{Evaluation}

Consider a \textbf{padded} reference-hypothesis target pair.
We identify the split between the result segment $R$
and the padding segment $P$ of the reference by the first occurrence of $c$, as illustrated in
\cref{fig:refhyp}. 
The corresponding segments in the hypothesis are indicated by an additional tilde:
$\tilde R, \tilde P$.
For the \textbf{simple} tasks, the padding is always the
empty string $\epsilon$ and is not considered further.
We then define a set of error predicates, see \cref{tab:predicates}.
The given instance-wise error is calculated as the indicator function of
the corresponding predicate
(possibly restricted only to $R,\tilde R$ for the unpadded, \textbf{simple}, tasks):
\begin{align*}
    \text{err}_\phi(RP, \tilde R \tilde P) = \begin{cases}
        1, & (RP, \tilde R \tilde P) \models \phi \\
        0, & (RP, \tilde R \tilde P) \not\models \phi.
    \end{cases}
\end{align*}
That is, the indicator is defined as $1$ if its predicate $\phi$ holds for the
given reference-hypothesis pair and $0$ otherwise.
Finally, this error indicator is aggregated over the validation dataset by averaging. Evaluation
was run every 20 training epochs.

\section{Results}

In general, the multi-task models (\textbf{$^*$/all})  achieved the same or 
worse performance than the corresponding specialized
models. The only substantial difference was the generalization to shorter sequences
for \textbf{copy} and \textbf{flip}. From
\cref{fig:error-any} plot of ``Error -- result any'', 
unlike fully specialized models, the
general ones attained a very high error on shorter sequences in 
\textbf{simple} tasks (left). For example,
while 
the ``Error -- result any'' indicator curve for 
\textbf{copy} is a flat $0$ over short sequences, it is $1$
for \textbf{copy/all}, and similarly
for \textbf{flip} and \textbf{flip/all}.
This discrepancy mostly disappeared in the respective
\textbf{copy/all}, \textbf{flip/all padded} tasks with only
a local error spike right under the minimum training length threshold,
see the rapid increase and decrease of the red and brown lines in the plot 
``Error -- result any'' for \textbf{padded} tasks (right) at length
values around $28$.
Training on different length ranges resulted only in a respective error shift 
towards the new training range without any significant difference in the overall
performance (see \cref{fig:range_comparison} in \cref{app:results}). 
For specialized models,
\textbf{simple} and \textbf{padded} variants also behaved similarly. 

\subsection{Generalization}

The largest differences in the behaviour were observed between \textbf{reverse} and the \textit{elementwise}
functions \textbf{copy}, \textbf{flip}.
The errors in the result strings
increased always immediately as argument length surpassed the training limit,
see
\cref{fig:error-any} ``Error -- result any'' for both \textbf{simple} (left) and 
\textbf{padded} (right) tasks.
On the other hand, the error was often nearly or fully absent for
sequences \emph{shorter} than the training range. This is the case for the tested elementwise functions 
(but not \textbf{reverse}). Further, for elementwise functions, the generation
for longer arguments
fails predominantly due to premature termination 
(\cref{fig:error-any} ``Error -- result short'')
while the generated prefix is mostly completely correct 
(\cref{fig:error-any} ``Error -- result prefix/long/short'').
We observe also a similar behaviour in the complementary case where the model was
trained on strictly longer inputs than the test cases.
The exact match test evaluates the result as wrong 
(\cref{fig:error-any} ``Error -- result any'')
but the common prefix between the (shorter) reference and the hypothesis is 
actually correct: 
\cref{fig:error-any} ``Error -- result prefix``, all elementwise tasks, i.e.
blue, green, red, and brown lines. Particularly,
the blue and green lines in \cref{fig:error-any} ``Error -- result prefix''
are at zero along the entire validation length range.
What follows in such excessively long hypotheses are some random $(a|b)^*$ tokens.
Usually the last valid token
according to the reference was repeated a few times.

\subsection{Indicator trajectories}

We tracked the trajectories of the individual error indicators over the entire
training process. The results vary across predicates and tasks. 
For example, \textbf{reverse} has never obtained a generalization ability that would be lost
later due to overtraining. Instead, its error curve is a gradually deepening U-shape,
as illustrated in \cref{fig:trajectories},
documenting that it slowly learns to carry out the reverse task within 
the training data range.
Contrarily, the elementwise functions usually generalize to shorter sequences,
even if imperfectly, by longer training, as shown for
\textbf{copy - simple} in \cref{fig:trajectories}.
Interestingly, this ability to operate with shorter sequences was mostly gained 
\emph{after} the performance on the training range plateaued 
(for a detailed illustration, see \cref{fig:late_descent} in \cref{app:results}).
The opposite behaviour, as reflected in the learning trajectory,
has been observed repeatedly for padding shortness
in the elementwise function tasks,
as illustrated for \textbf{copy - padded}
(\cref{fig:trajectories} ``Error -- padding short'').
In such cases, models for elementwise functions often started to unexpectedly diverge from 
the training invariant of exactly 70-token target length
(with an appropriate $(a|b)^*$ result segment and a subsequent $cd^*$ padding to the
total constant length of 70 tokens, as shown in \cref{fig:example}) and 
terminated the padding (and entire hypothesis) generation prematurely.

\subsection{Padding shortness}

\begin{figure}[b]
    \includegraphics[width=0.45\textwidth]{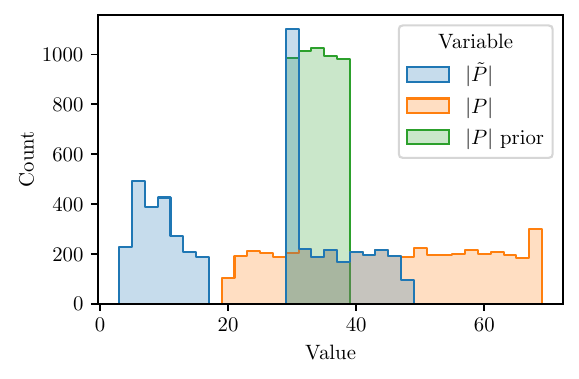}

    \caption{\textbf{Copy - padded} distributions 
        (\textit{Count} - total number of validation examples 
        with a given value \textit{Value}) for
        hypothesis padding lengths $|\tilde P|$ (\csquare[blue!50] blue),
        reference padding lengths $|P|$ (\csquare[orange!50] orange), 
        and the training padding length $|P|$ prior (\csquare[green!50] green).
    }
    \label{fig:distributions}

\end{figure}

\begin{figure}[t]
    \includegraphics[width=0.49\textwidth]{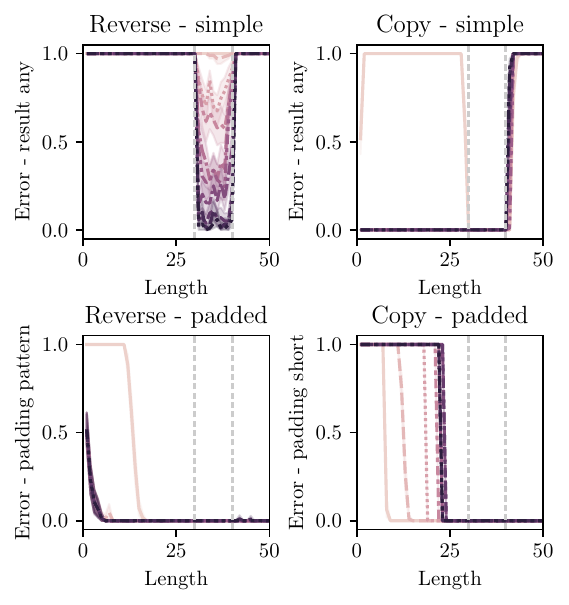}
    \caption{Selected indicator trajectories throughout training. 
        40 epochs (\csquare[orange!30] light orange) to
        400 epochs (\csquare[violet!80] dark violet).}
    \label{fig:trajectories}
\end{figure}

Padding shortness is a phenomenon in which models trained on \textbf{padded} tasks
do not correctly pad the
$(a|b)^*$ result segment by the subsequent $cd^*$ padding segment to the
total length of 70 tokens as trained (see \cref{fig:example}),
but terminate the padding (and hypothesis) generation prematurely.
We observed padding shortness in all the considered elementwise tasks.
The phenomenon occured more frequently during later epochs. 
In particular, padding shortness
occured initially only for very short argument lengths and gradually became a problem
for longer arguments, closer to the training argument length range
(\cref{fig:trajectories} ``Error -- padding short'').
\cref{fig:distributions} illustrates the unwillingness of the Transformer model
to generate padding lengths significantly exceeding the training distribution:
compared with the validation reference distribution of padding lengths, the model
hypotheses
remain greatly influenced by the training padding length distribution, leading
to this phenomenon.
Moreover, the padding generation has been found to sometimes ``short circuit'' terminate
for very short arguments, whereby the \texttt{EOS} token is emitted (after the correct
result is generated) by the
model even before any familiar padding length is reached (5-15 padding tokens 
in \cref{fig:distributions}),
resulting in even shorter 
overall hypotheses (e.g. 10-25 tokens of result and padding in total).

\section{Discussion}

\subsection{Longer sequences}

We have found that our models universally struggle with generalization
to arguments longer than the training limit,
i.e. longer than what was demonstrated in the training data.
This failure is \textit{immediate
and complete}, with error rates
quickly approaching $1$ (\cref{fig:error-any}). 
Particularly, the models were not willing to generate the $(a|b)^*$ results beyond the training
length limit, terminating prematurely with \texttt{EOS} in \textbf{simple} tasks and 
$cd^*$ in \textbf{padded} tasks.

For the elementwise functions studied (\textbf{copy},
\textbf{flip}), the partial results the models generated were however
correct prefixes of the references. This suggests the models 
\textit{do not hallucinate, but are unwilling}. 
It may be therefore possible the models have gained an implicit generalization ability
which could be revealed by forcing them to continue the generation
and inspecting the
relative probabilities of the tokens \cite{distilling} by masking the \texttt{EOS}, \texttt{PAD}
tokens. A more easily implementable solution could be masking out the $c,d$ tokens 
found in
\textbf{padded} tasks.
Masking the $c,d$ tokens can be achieved by possibly only
using out-of-the-box arguments such as \textit{bad words} in many frameworks,
avoiding manual modification of the self-regression loop code, which
would be otherwise necessary in order to force the models to continue generating.

In case of \textbf{reverse} requested from longer inputs than the training data,
the prefixes were incorrect.
However the prefix analysis may not be sufficient when analyzing the extent to 
which \textbf{reverse} hypotheses match the references,
as the task is based on permutation, instead of a left-right parsing and generation.
Metrics that perform a more sophisticated matching, such as BLEU \cite{bleu}, could be used 
here to test the extent to which the partial \textbf{reverse} hypotheses are 
related to the desired solution.

\subsection{Shorter sequences}

Generalization to shorter arguments has been observed for
all the elementwise functions. This ability was not always observed with models
trained on 
all the string edit functions simultaneously.
Because the sizes of our specialized Transformer models were already near the 
theoretical lower bounds for the individual functions obtained
by RASP \cite{thinking}, it may be possible that the models
trained in the multi-task setting on all the functions simultaneously
become under-parameterized and forced into various trade-offs,
which may even allow errors that usually e.g. occur only for \textbf{reverse} to 
\textit{pervade}
into other tasks. 
This might be compensated by increasing the model size accordingly,
e.g. linearly scaling the number of attention heads with the number of functions,
but we leave such experiments for the future.




We note that our experiments with the \textbf{reverse} 
function were limited to 1-layer Transformer while \citet{thinking} indicate
that the minimum number of layers should be 2. 
For now, we know that on the independent and identically distributed 
validation subset with a familiar
length range, our models achieved perfect or near-perfect performance, i.e. 
\textbf{reverse} in a known restricted range is learnable even with a smaller model.
We are now running the missing experiments with a deeper model for 
\textbf{reverse} to find out if the deeper models allow Transformer 
to generalize \textbf{reverse} at least to shorter lenghts, or if it will fail. 
It may be possible that even considerably larger models 
will struggle with generalizing \textbf{reverse}, 
because the attention matrices
needed to correctly reverse sequences of different lengths may be highly dependent
on these lengths and not easily learned from training data with a restricted length
range.

The occurrence of generalization to shorter arguments is interesting, 
esp. in the conjunction with the inability (or unwillingness) observed for longer arguments.
The models correctly generate the results for shorter arguments and terminate
them with \texttt{EOS} or $cd^*$ at positions never seen during training.
Together with the prefix-correctness observed for longer arguments, this
demonstrates that Transformer is able to generalize to lengths unseen
during training in some scenarios, while possibly struggling in others
\cite{overfitting,varis}. 
Further, this suggests that Transformer is able to learn and generalize 
algorithmic aspects of the 
string edit tasks. This may be constrained by the size of the model or the complexity of
the task, with the possible distinction between \textit{elementwise} tasks,
where the model can only attend to the matching tokens in the source and
target sequences, and tasks with a more complex token pairing, such as \textbf{reverse}.

The generalization to shorter arguments frequently occurred 
mostly
after the performance
on the training length range plateaued. We attribute this to the stochasticity
of the SGD, which may allow further re-parameterization of the trained models
based on the nonzero and
noisy loss obtained from permutations of the training data.
SGD is moreover hypothesized to facilitate double descent, where model 
over-parameterization
together with longer training eventually leads to generalization and improvement on the validation data after the training set has
been already fitted \cite{double}. In other words, SGD seems to lead to models that generalize 
surprisingly well with seemingly no motivating signal from the training data.
Further, Dropout
\cite{dropout}
introduces implicit ensemble learning, whereby a slightly different
network is present during each training iteration based on the randomized masking
of the artificial neurons by the Dropout \cite{ensemble}, which can also
contribute to the late re-parameterization of the whole network.

\subsection{Padding shortness}

While the result error on shorter sequences disappeared, the phenomenon of
padding shortness emerged. We propose this shows an antagonism between
different training data priors the model is simultaneously trying to adhere to.
Namely, a hypothesis consists of the result $\tilde R$ and the padding
$\tilde P$. A model trained to generate $|\tilde R| \in (30, 40]$ and padded to 
$|\tilde R \tilde P| = 70$ tokens is therefore 
implicitly trained with 
$|\tilde P| = 70 - |\tilde R| \in [30, 40)$.
When the model subsequently generates a result of length e.g. $|\tilde R| = 10$
during adversary testing, it cannot satisfy both the padding length prior 
$|\tilde P| \in [30, 40)$
and the total hypothesis length prior $|\tilde R \tilde P| = 70$.
The longest familiar
padding $|\tilde P| = 39$ then leads to an insufficient overall length of
$|\tilde R \tilde P| = 10 + 39 = 49 < 70$ and vice versa.
That is, shorter-than-expected arguments lead to
the need of longer-than-expected paddings.

The model is aware of multiple patterns in the training data. This may include
statistics concerning the structure of the training sequences,
such as the distribution of the \texttt{EOS} token or other token positions, 
segment lengths, or spurious co-occurrences, and
also the inner algorithmic or language modelling aspects in question.
During testing on out-of-distribution data, these patterns may come into conflict,
whereby the Transformer model often favours the different structural statistics
(adhering to the padding length $|\tilde P| \in [30, 40)$, 
result length $|\tilde R| \in (30, 40]$,
total length $|\tilde R \tilde P| = 70$), while still
aware of the inner modelling aspects
(copying, flipping).
This preference may moreover change
during training, as the model is gradually re-parameterized. Viewed as
ensemble learning \cite{ensemble}, Dropout might facilitate such a re-parametrization
by the 
simultaneous training
of multiple implicit subnetworks with mutually different preferences.

We see the idea of adversarial testing and some inclusion of a novel form 
of adversarial training as a possible future path to models that, 
when forced to generalize beyond the training data domain, adhere to those 
of conflicting properties that we would prefer them to.
Length overfitting remains problematic, as it may occur at unexpected situations, 
especially in more complex natural language settings \cite{overfitting,varis}.
The models used in natural language modelling and natural language 
translation are moreover considerably larger and have access to much richer
training data, which may not solve this problem alone \cite{overfitting,
varis}, and
in such scenarios, partial correctness of the results is often not
satisfactory.
Modifications such as 
relative position representation \cite{relative_encoding} may help Transformer
to generalize in many tasks such as the string edit functions considered.


\section{Conclusion}

Using adversary testing, we demonstrate that the Transformer model can generalize
only partially with regard to the length of 
the training sequences in elementary string edit tasks.
The largest differences emerged between the elementwise functions considered 
(copying, swapping $a$ for $b$ and vice versa) and string reversing.
Generalization to shorter sequences occurs systematically for certain variants
of the considered
elementwise functions but never for string reversing. 
On the other hand, Transformer
was found to immediately struggle with generalization to longer sequences across 
all tasks considered, agreeing with previous research on the topic.

Beyond the extent of previous research, we show that even in the failing cases,
the models do not hallucinate and still
adhere to the tasks, subsequently interrupted by prematurely emitting a terminating
token. In particular, the hypotheses generated were often completely 
valid prefixes of the references.
We find evidence the models are aware of various characteristics of the training data,
including the algorithmic aspects of the tasks, while possibly favouring the
undesired ones in situations when they come into conflict. 
We suggest that exploring these learned characteristics and their conflicts 
arising in adversary tests could be a good inspiration for future training 
approaches, aiming at controllable preferences when a conflict happens.

From the practical point of
view, we confirm that controlling for sequence length and other structural 
properties in the training
data is an important step. Transformer is influenced by these characteristics
and the obtained models may as a result behave unexpectedly with new data.



\section*{Impact statement}

The goal of our work is to advance the field of Machine Learning. 
There are many potential societal consequences of our work
(esp. in the application of Transformer models in Large Language Models), 
none which we feel must be specifically highlighted here.

\bibliographystyle{icml2024}
\bibliography{biblio.bib}

\clearpage

\appendix

\section{Model details}
\label{app:models}

All models were implemented in Fairseq\footurl{https://github.com/facebookresearch/fairseq}
\cite{fairseq} with the hyperparameters 
reported in \cref{tab:hyperparams}. The format of training examples follows \cref{fig:example}.

\begin{table}[h]
    \centering
    \caption{Model hyperparameters}
    \label{tab:hyperparams}
    \begin{tabular}{l l}
        \toprule
        Hyperparameter & Value \\
        \midrule
        Embedding size & $128$ \\
        Feedforward size & $512$ \\
        Depth & $1$ \\
        Attention heads & $1$ \\
        Learning rate & $10^{-4}$ \\
        Training epochs & $400$ \\
        Batch size & $4096$ tokens \\
        Dropout & $0.2$ \\
        Warmup & $4000$ steps \\
        Label smoothing & $0.1$ \\
        \bottomrule
    \end{tabular}
\end{table}

The training and evaluation times for a single model ranged 6-36 h, depending
on the task 
(\textbf{$^*$/all - padded} tasks taking the longest). Our source code will be made publicly available upon acceptance.%

\section{Results details}
\label{app:results}
\enlargethispage{1cm} 
\begin{figure}[h!]
    \centering
    \includegraphics[width=0.49\textwidth]{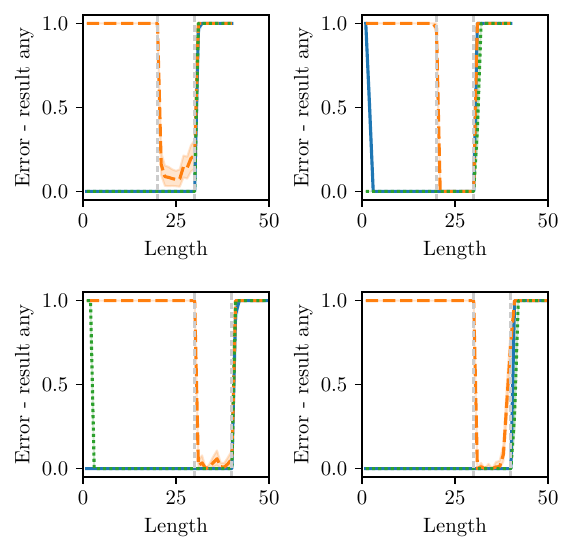}
    \caption{Selected indicators after 400 epochs. \\
             Left - \textbf{simple} tasks, right - \textbf{padded} tasks.\\
        \textbf{Copy} (\csquare[blue!60] blue), 
        \textbf{flip} (\csquare[green!60] green),
        \textbf{reverse} (\csquare[orange!60] orange). \\
        Training length range indicated by vertical dashed lines \\
        (top row - $(20, 30]$, bottom row - $(30, 40]$).
    }
    \label{fig:range_comparison}
\end{figure}

\begin{figure}[t]
    \centering
    \includegraphics[width=0.49\textwidth]{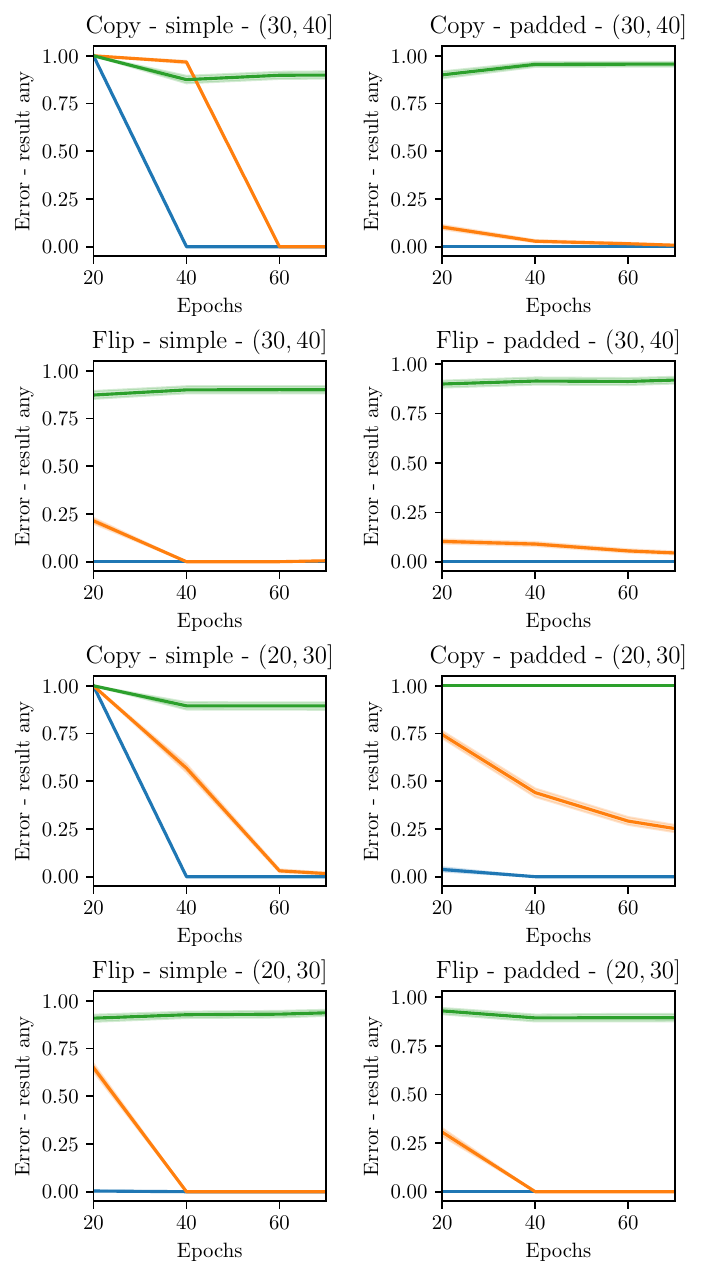}
    \caption{\textit{``Function - variant - training length range''} trajectories of the
        \textit{result any} indicator evaluated on
        validation subsets based on the argument length:
        Training length range (either $(20, 30]$ or $(30, 40]$; \csquare[blue!60] blue),
        lower lengths (\csquare[orange!60] orange), 
        higher lengths (\csquare[green!60] green).}
    \label{fig:late_descent}
\end{figure}


\end{document}